\documentclass{article}

\usepackage{arxiv}
\usepackage{subfigure}
\usepackage{adjustbox}
\usepackage{lineno,hyperref}
\modulolinenumbers[5]
\usepackage{algpseudocode}
\usepackage{graphicx}
\usepackage[most]{tcolorbox}

\usepackage{amsmath,amssymb,amsfonts}
\usepackage{amsthm}
\usepackage{booktabs}
\theoremstyle{definition}

\usepackage{url}
\usepackage{enumitem}
\setlist{leftmargin=5mm}
\usepackage{xcolor}
\usepackage[ruled]{algorithm2e}

\SetCommentSty{colortcp}
\usepackage{mathtools}
\DeclarePairedDelimiterX{\infdivx}[2]{(}{)}{#1\;\delimsize\|\;#2}

\usepackage{color}

\title{Seed Stocking Via Multi-Task Learning}

\author{
 Yunhe Feng \\
  Electrical Engineering \& Computer Science\\
  University of Tennessee\\
  Knoxville, TN 37996 \\
  \texttt{yunhefeng@utk.edu} \\
  \And
  Wenjun Zhou \\
  Business Analytics \& Statistics\\
  University of Tennessee\\
  Knoxville, TN 37996 \\
  \texttt{wzhou4@utk.edu} \\
}

\begin{document}

\maketitle

\begin{abstract}
Sellers of crop seeds need to plan for the variety and quantity of seeds to stock at least a year in advance.
There are a large number of seed varieties of one crop, and each can perform best under different growing conditions.
Given the unpredictability of weather, farmers need to make decisions that balance high yield and low risk.
A seed vendor needs to be able to anticipate the needs of farmers and have them ready.
In this study, we propose an analytical framework for estimating seed demand with three major steps.
First, we will estimate the yield and risk of each variety as if they were planted at each location.
Since past experiments performed with different seed varieties are highly unbalanced across varieties,
and the combination of growing conditions is sparse,
we employ multi-task learning to borrow information from similar varieties.
Second, we will determine the best mix of seeds for each location by seeking a tradeoff between yield and risk.
Third, we will aggregate such mix and pick the top five varieties to re-balance the yield and risk for each growing location.
We find that multi-task learning provides a viable solution for yield prediction, and our overall analytical framework has resulted in a good performance. 
\end{abstract}

\keywords{modern portfolio theory \and mean-regularized multi-task learning \and graph based multi-task learning \and crop yield prediction \and seed selection}

\section{Introduction}
% background & problem important?
With the rapid growth of the global population, food demand increases proportionately.
The food supply crisis is becoming a social and global problem.
Based on a recent survey by the Food and Agriculture Organization (FAO), food production will need to rise by 70\% by 2050 when the population reaches 9.1 billion, leading to severe food reduction and food shortage.
Food production is also negatively affected by the degradation of Earth’s ecosystems and more frequent adverse weather like El Niño, as exemplified by the 2011 East Africa drought and the 2010 Sahel famine.
One straightforward and possible solution for this problem is to plant crop seed varieties that can simultaneously maximise crop yields and maintain strong resistance to poor growing conditions.
Therefore, selecting and stocking appropriate seed varieties for a large farmland area with varied climate and soil conditions becomes crucial, especially for seed retailers.

% challenge ?
However, the selection of optimal seed combinations for stocking is very challenging for the following reasons.
First, a large number of available seed varieties make it difficult to distinguish their characteristics and evaluate their performances in different growing conditions.
Second, many external factors, like weather and soil conditions, may affect agricultural land productivity even if seed varieties have been determined.
Third, growing conditions where the seeds will be planted have to be predicted one year in advance.

% existing work? research gap?
To improve crop yields, most existing works either focus on how to improve certain seed internal features or explore which external factors may affect agricultural output.
Kasuga et al.~\cite{kasuga1999improving} adopted gene transfer techniques to improve seed performance in drought, salt, and freezing stress.
Teramura et al.~\cite{teramura1990effects} studied the effects of UV-B radiation on crop yields and seed quality.
Brevedan and Egli~\cite{brevedan2003short} illustrated the relationship between a short period of water stress and agricultural productivity.
Samarah et al.~\cite{samarah2005effects} explained the effects of drought stress on growth and yield of barley.
Few works investigate how to select seed combinations for stocking to optimize the future yield of a large area of farmland with various growing conditions.
We think the lack of relevant data, which is under the protection of business secrets or patents, may mainly cause such a research gap.

% our approach
In this paper, we propose an analytical framework to select soybean seed candidates with high yields and low risks for stocking.
Multi-task learning techniques are applied to predict crop yields of different seed varieties under diverse weather and soil conditions.
In particular, we treat the planting of each soybean variety as an individual task, which takes multiple variables including climate, soil conditions, planting date and product risks as inputs and the yield as output.
On the one hand, all these soybean varieties share some latent commonalities because they share common ancestors and have a high level of soybean-specific genetic similarity.
On the other hand, different seed varieties might have exclusive and unique characteristics, for example, strong drought resistance and salt tolerance.
The commonalities are captured while doing the induction of multiple tasks, while the uniqueness is represented by the individual parameter for each task.
The multi-task regularization provides a mechanism for borrowing information from similar training records because it forces these tasks to be learned together.
To the best of our knowledge, this is the first work that adopts multi-task learning to select seeds for future seed stocking.

To strike a balance between yield and risk, we utilize modern portfolio theory (MPT) to select optimal and sub-optimal seed varieties by considering the tolerance risk of each farming location
Based on the seed varieties selected in the above step, we derive the global optimal seed variety combinations that yield maximum productivity and minimize the uncertainty and risks associated with growing conditions.

The rest of the paper is organized as follows.
We will first review the related work in Section \ref{related},
and then we will describe the datasets and research problem in Section \ref{dataset}.
Our methods will be described in Section \ref{framework},
followed by implementation details in Section \ref{design}.
Section \ref{evaluation} presents the results and our evaluation of effectiveness.
Finally, we will draw conclusions and discuss future work in Section \ref{conclusion}.

\section{Related Work}\label{related}

We categorize the related work into two parts.
First, we introduce the existing literature exploring factors that influence crop yields.
Then we briefly introduce the development of multi-task learning and its applications. 

\subsection{Factors in Crop Production}
Many works have studied factors that affect crop yields. We briefly summarize factors of internal seed genetic characteristics and external planting conditions in the following subsections.

\subsubsection{Internal Genetic Factors}

For internal genetic factors,
Seker and Serin~\cite{seker2004explanation} explained the relationships between seed yield and some morphological traits in smooth bromegrass using path analysis.
Yin and Vyn~\cite{yin2005relationships} illustrated the relationship of isoflavone, oil, and protein in the seed with crop yield.
Ghaderi et al.~\cite{ghaderi1984relationship} revealed the relationship between genetic distance and heterosis for yield and morphological traits in dry edible bean and faba bean.
{Bola{\~n}os-Aguilar et al.~\cite{bolanos2000genetic} summarized the genetic variation for seed yield and its components in alfalfa (Medicago sativa L.) populations.
Rao et al.~\cite{rao2008genetic} conducted a comprehensive study of genetic associations, variability and diversity in seed characteristics, growth, reproductive phenology and yield in Jatropha curcas (L.) accessions.
Specht et al.~\cite{specht1999soybean} presented crop yield potential from a genetic and physiological perspective.
Kasuga et al.~\cite{kasuga1999improving} adopted gene transfer techniques to improve seed performance in drought, salt, and freezing stress to increase plant productivity.

\subsubsection{External Environmental Factors}

Many environmental variables, such as weather and soil conditions, may affect crop yields.
To be specific, the weather conditions include temperature, precipitation, solar radiation and so on.
The soil conditions contain H+ concentration, soil texture, organic matter, Cation Exchange Capacity (CEC) and so on.

\textbf{Temperature.}
From the perspective of temperature, Schlenker and Roberts~\cite{schlenker2009nonlinear} reported nonlinear temperature effects indicated severe damages to US crop yields under climate change.
Wheeler et al.~\cite{wheeler1996growth} studied the growth and yield of winter wheat (Triticum aestivum) crops in response to $CO_2$ and temperature.
Lobell and Field~\cite{lobell2007global} revealed the relationship between global warming and crop yields.

\textbf{Precipitation.}
Samarah~\cite{samarah2005effects} investigated the effects of drought stress on growth and yield of barley.
Musick et al.~\cite{musick1994water} demonstrated the water-yield relationships for irrigated and dryland wheat in the US Southern Plains.
Brevedan and Egli~\cite{brevedan2003short} illustrated the relationship between a short period of water stress and crop yield.
Wheeler et al.~\cite{wheeler2000temperature} reviewed the evidence for the importance of variability in temperature for annual crop yields.

\textbf{Solar Radiation.}
Hipp et al.~\cite{hipp1970influence} studied the influence of solar radiation and planting date on yield of sweet sorghum.
Teramura et al.~\cite{teramura1990effects} explored the effects of UV-B radiation on crop yield and seed quality.
Amir and Sinclair~\cite{amir1991model} proposed a model of how temperature and solar-radiation affect spring wheat growth and yield.
Muchow et al.~\cite{muchow1990temperature} utilized a simple, mechanistic crop growth model to examine the effects of variation in solar radiation and temperature on potential maize (Zea mays L.) yield among locations.

\textbf{Soil Texture.}
Sene et al.~\cite{sene1985relationships} revealed the relationships of soil texture and structure to corn yield response to subsoiling.
Cox et al.~\cite{cox2003variability} aimed to determine the variability of selected soil properties and the relationship between these soil properties and crop yield.
Insam et al.~\cite{insam1991relationship} studied the relationship of soil microbial biomass and activity with fertilization practice and crop yield in three ultisols.
Bhogal et al.~\cite{bhogal2003effects} investigated the effect of past sewage sludge additions on heavy metal availability in light textured soils, and presented its implications for crop yields and metal uptakes.

\textbf{Soil Organic Matter.}
Griffith et al.~\cite{griffith1988long} presented long-term tillage and rotation effects on corn growth and yield in poorly drained soils with both high and low organic matter content.
Soane~\cite{soane1990role} reviewed the role of organic matter in soil compatibility from several practical aspects.
Pan et al.~\cite{pan2009role} studied the role of soil organic matter in maintaining the productivity and yield stability of cereals in China.
Diaz-Zorita et al.~\cite{diaz1999soil} explored the relationship between soil organic matter and plant productivity but in the semiarid Argentine Pampas.

\subsection{Multi-task Learning}

Multi-task learning techniques learn one task together with other related tasks simultaneously by taking advantages of the commonality among the tasks.
To utilize multi-task learning, the assumption that all or some of the involved tasks share related similarities must be satisfied.
Many variations of multi-task learning algorithms have been proposed.
Evgeniou and Pontil~\cite{evgeniou2004regularized} proposed regularized multi-task learning which assumed the task parameter vectors of all tasks are close to each other.
Argyriou et al.~\cite{Argyriou06} introduced multi-task learning with joint feature learning to capture the set of features representing the relatedness of tasks.
Jalali et al.~\cite{jalali2010dirty} developed a dirty model for multi-task learning to deal with dirty data that might not fall into a single structure.
To process outlier tasks, Chen et al.~\cite{chen2011integrating} designed robust multi-task learning by using the low-rank structure to capture task relatedness and a group-sparse structure to identify the irrelevant tasks.
Zhou et al.~\cite{zhou2011clustered} proposed a clustered multi-task learning formulation based on the spectral relaxed $k$-means clustering.
Multi-task learning techniques have been widely used in various field, such as biology \cite{xu2011multitask,liu2010multi,caruana2003benefitting}, finance \cite{Ghosn96}, chemistry \cite{unterthiner2015toxicity} and medicine \cite{zhou2011multi,zhang2012multi,bickel2008multi}.

\section{Data Description}\label{dataset}

This section describes two datasets, named ``Experiment'' and ``Region'', which are provided by one of the largest international crop seed companies.
We summarize the two datasets' attributes into four categories: coordinates, weather, soil conditions, and planting features.

\subsection{Dataset Overview}

\begin{itemize}
  \item
  The ``Experiment'' dataset represents the current knowledge on how the available seed varieties of one crop perform in different weather and soil conditions.
  It contains more than 82,000 records involving 174 varieties, planted in 583 locations, between 2009 and 2015.
  \item
  The ``Region'' dataset represents long-term knowledge of a large band of the U.S. Midwest with a similar growing season length. This is the region where the farmers' choice of seed (or proportion of different seeds) will be planted. It contains 6,490 locations and a record of growing conditions between 2001 and 2015.
\end{itemize}

\subsection{Feature Variables}

\begin{figure}[h]
\centering
\subfigure[Experiment\label{fig:1}]{\includegraphics[width=0.48\linewidth]{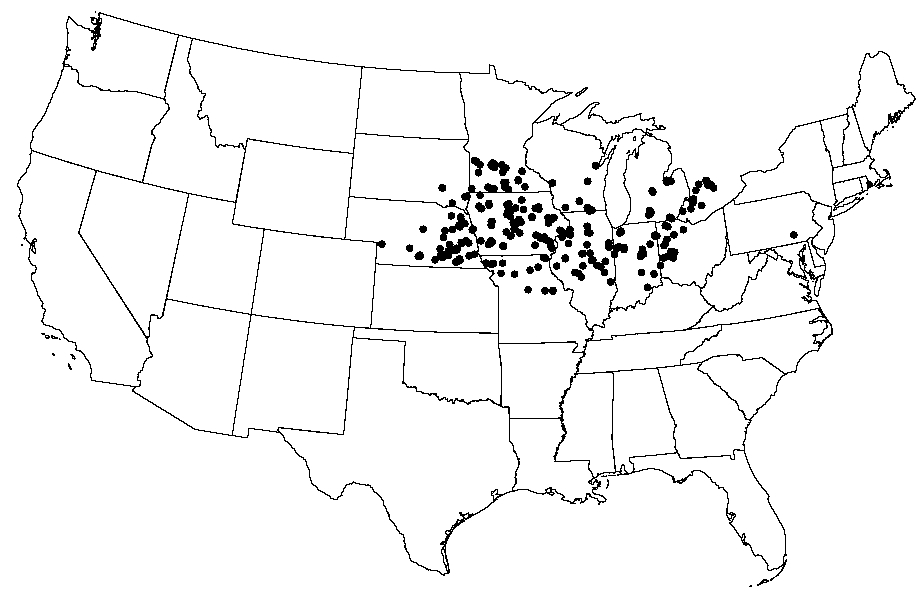}}
\hfill
\subfigure[Region\label{fig:2}]{\includegraphics[width=0.48\linewidth]{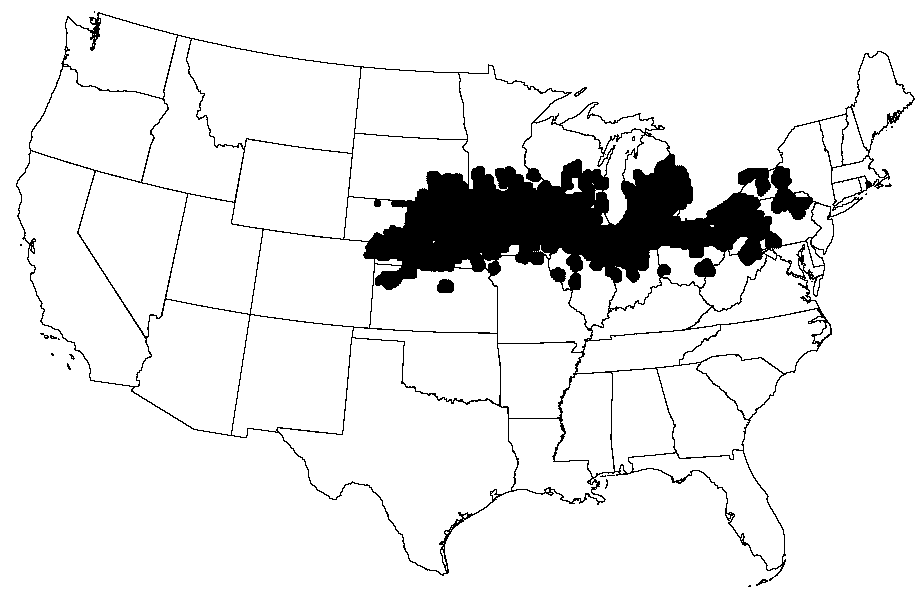}}
\caption{Locations in ``Experiment'' and ``Region'' datasets}
\label{fig:1and2}
\end{figure}

\begin{table}[ht]
\centering
%\scriptsize
\caption{Attributes in the datasets}\label{tab:1}
\begin{tabular}{lllcc} 
\toprule
\textbf{Category} & \textbf{Attributes} & \textbf{Meaning} & \textbf{Experiment} & \textbf{Region} \\
\midrule
Coordinates
& Year         & The year when the data are collected & 2009--2015    & 2001--2015 \\
& Lat. \& Lon. & Geo-coordinates of farmlands         & 583 locations & 6490 locations \\
\midrule
Weather
& Temperature     & Sum of daily temperatures    & \multicolumn{2}{c}{}\\
& Precipitation   & Sum of daily precipitation   & \multicolumn{2}{c}{varies by year and location}\\
& Solar Radiation & Sum of daily solar radiation & \multicolumn{2}{c}{}\\
\midrule
Soil
& CEC            & Cation Exchange Capacity (cmol kg-1)        &\multicolumn{2}{c}{}\\
& pH             & Log of H+ concentration in the soil         &\multicolumn{2}{c}{}\\
& Organic Matter & The percentage of organic matter in soil    &\multicolumn{2}{c}{varies by location only;}\\
& Soil Clay      & The percentages of  soil small particles    &\multicolumn{2}{c}{does not change by year}\\
& Silt           & The percentages of soil medium particles    &\multicolumn{2}{c}{}\\
& Sand           & The percentages of soil large particles     &\multicolumn{2}{c}{}\\
& PI$^{*}$       & The degree of suitability for growing crops &\multicolumn{2}{c}{}\\
\midrule
Planting
& Variety             & seed variety to be evaluated          & 174 varieties &\\
& Planting Date$^{*}$ & what day when the variety was planted & May-Sep.      & TBD\\
& Yield               & crop productivity                     &               &\\
\bottomrule
\end{tabular}
\begin{flushleft}
$^{*}$ has missing values
\end{flushleft}
\end{table}

The available variables in the two datasets are listed in Table~\ref{tab:1}.
We categorized these variables into four groups:
\begin{itemize}
  \item \textbf{Coordinates}: spatiotemporal coordinates of the farmland.
The spatial dimensions are represented with latitude (Lat.) and longitude (Lon.) pairs.
Figure~\ref{fig:1and2} visualizes the available locations of farmlands in the ``Experiment'' and ``Region'' datasets.
Locations in the ``Region'' dataset are much denser than in the ``Experiment'' dataset.
The temporal dimension is identified to the granularity of the year.
This is a reasonable granularity because farming is oftentimes planned and implemented by year.
The ``Experiment'' dataset includes data from 2009 to 2015, whereas the ``Region'' dataset covers a longer period of time, starting in 2001 and ending in 2015.

  \item \textbf{Soil}: the soil attributes, including
              the Cation Exchange Capacity (CEC, measured in cmol kg-1),
              the log of H+ concentration (pH),
              the percentage of the soil consisting of organic matter,
              percentages of small, medium, and large soil particles (Soil Clay, Silt and Sand),
              and the productivity index (PI), classified according to \cite{Hengl16}.
              These variables are fixed in a given location, and do not change over time.
              Therefore, when planning for the whole region, the soil conditions are treated as known.

  \item \textbf{Weather}: the weather condition variables, consisting of
                 total daily temperature, precipitation, and solar radiation,
                 as aggregated per year between April 1st and October 31st.
                 These quantities vary by both location and year.
                 Therefore, in order to plan for the next year, the weather conditions are uncertain.
                 Since weather condition for the next year is generally unpredictable,
                 we should consider these as random variables for future years.

  \item \textbf{Planting}: the planting decisions and outcome variables in the ``Experiment'' dataset.
                 These include the seed variety planted, the planting date, and the yield information based on prior years' experimentation.
                 They are to be determined for the ``Region'' dataset.
                 Specifically, our task is to determine composition of seed varieties per location to maximize yield given uncertain weather conditions.
                 We have ignored the planting date in this study, instead assuming that the selected seeds are planted during the right season on a reasonable day.
                 Even though the planting date is likely a relevant decision element,
                 we feel that it may have to do with scheduling constraints, for which we have very limited information.
\end{itemize}

\section{Method}\label{framework}

Generally, our goal is to select the seed variety combinations which will optimize the productivity of the farms in the ``Region'' dataset.
Such decisions will be made based on the yield models of different seed varieties learned from the ``Experiment'' dataset.
Therefore, we divide the work into three steps: estimation, projection, and planning, as outlined in Figure~\ref{fig:framework}.

\begin{figure}[h]
  \centering
  \includegraphics[width=\linewidth]{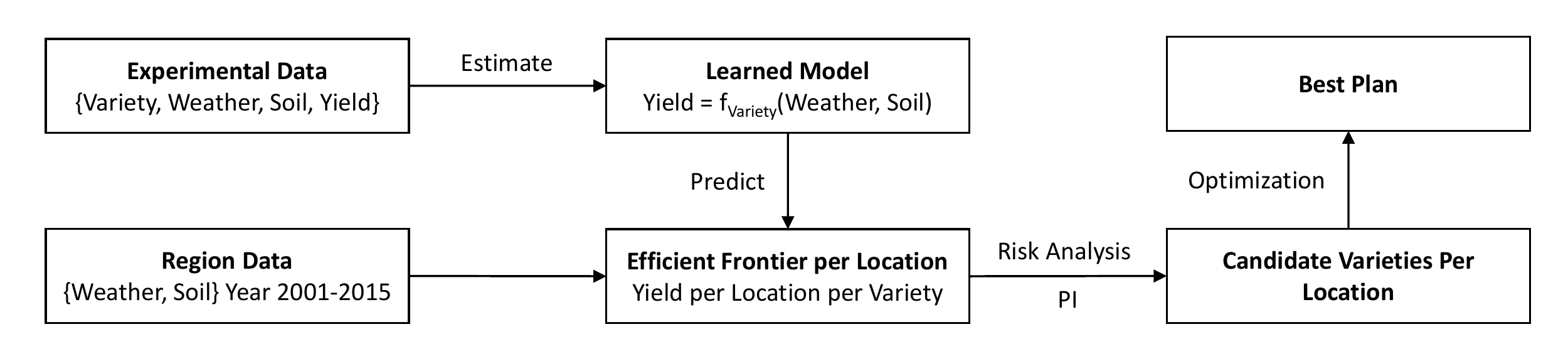}
  \caption{The framework of our solution}\label{fig:framework}
\end{figure}

\subsection{Estimation}

Our first step is to learn about seeds' yield distributions under each possible set of growth conditions using the ``Experiment'' data.
Formally, let $i$ be the index for seed variety,
we need to learn a function $f_i$ that maps its yield for given weather and social conditions:
\begin{equation}\label{}
  f_i: \mathcal{X} = (\mathcal{W}, \mathcal{S}) \rightarrow Y
\end{equation}
where $\mathcal{W}$ represents the joint distribution of weather, and $\mathcal{S}$ represents the combination of soil conditions.
Therefore, the first challenge we are facing is how to put all variables together to create a comprehensive model.
For simplicity, we will employ a linear model. 
Our training data could be derived from prior experimental data, which would provide the actual yield for various combinations of growing conditions.

However, the experimental data are extremely imbalanced in seed varieties chosen.
In other words, there are a small number of seed varieties with a lot of experimental data and many more seed varieties with very limited experimental data.
Note that the number of observations per seed variety in the ``Experiment'' dataset is highly skewed, as shown in Figure~\ref{fig:observation_distribution}.
Specifically, more than 75\% of varieties have less than 500 observations.

\begin{figure}[h]
\centering
\includegraphics[width=0.6\textwidth]{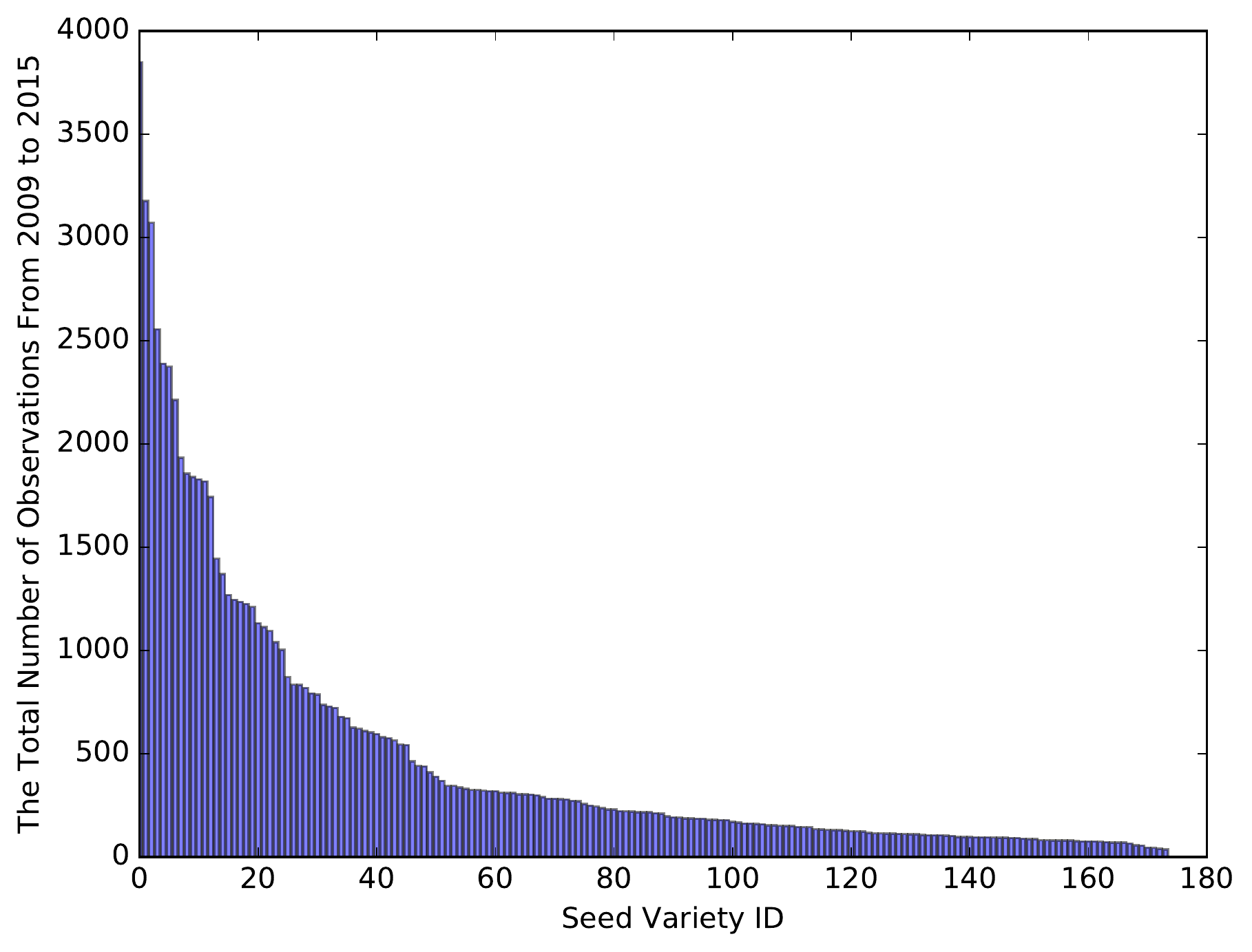}
\caption{Number of observations per seed variety}
\label{fig:observation_distribution}
\end{figure}

Despite just a dozen variables on growing conditions, the set of all possible combinations is prohibitively large, making the experimental data sparse for estimating yield under each combination.
To overcome the data sparsity problem, we are motivated to use multi-task learning, which provides a mechanism for borrowing information from similar training records.
In particular, we treat the planting of each seed variety as an individual task, which takes multiple variables including climate, soil conditions, planting date and product risk as inputs and the yield as output.
We believe all these seed varieties share some latent commonalities because they all belong to the same crop category, which means they share common ancestors and have a high level of crop-specific genetic similarity.
Moreover, different seed varieties have their own specific minor characteristics, such as strong drought resistance and salt tolerance.

There are at least two commonly used approaches for formulating multi-task learning: mean-regularized multi-task learning and graph based multi-task learning. In the following, we will describe each formulation in detail.

\subsubsection{Mean-Regularized Multi-Task Learning}

Mean-regularized multi-task learning assumes that all tasks (i.e., the planting of seed varieties) are close to the average, and may deviate from the mean task.
Suppose that for variety $i$, there were $n_i$ experiments done in the past, then the training data matrix for this variety, $X_i$, would have a size $n_i \times (p+1)$.
The columns are $p$ growing condition variables plus a constant.
Accordingly, $Y_i$ is a vector of length $n_i$, which consists of the actual yield of those $n_i$ experiments.
Then, the mean-regularized multi-task learning formulation can be expressed as follows:
\begin{eqnarray}\label{eq:1}
\min_{W} && \sum_{i=1}^V \begin{Vmatrix} X_i\vec{\beta}_i - Y_i \end{Vmatrix}_{F}^{2} +
\lambda \sum_{i=1}^V \begin{Vmatrix} \vec{\beta}_i - \frac{1}{V} \sum_{j=1}^V \vec{\beta}_j \end{Vmatrix}_1
\end{eqnarray}
where $\| \cdot \|_{F}^{2}$ represents the squared Frobenius norm,
$\| \cdot \|_{1}$ represents the $L_1$-norm,
$\lambda$ is a regularizing parameter for mean-regularization,
and $V$ is the total number of varieties possible.
Our goal is to find the best $W = \left[\vec{\beta}_1, \vec{\beta}_2, \dots, \vec{\beta}_V\right]$,
where $\vec{\beta}_i$ is a coefficient vector of length $(p+1)$, representing the (linear) model of task $i$.

\subsubsection{Graph Based Multi-Task Learning}

Considering that some seed varieties may share a stronger similarity in terms of growth characteristics (e.g., strong drought resistance or salt tolerance) than others, we can integrate more flexibility by incorporating an affinity graph.
This approach, in the context of multi-task learning, is known as graph based multi-task learning.
Specifically, suppose that $G$ is the graph structure (represented as a $V \times V$ matrix) incorporating the similarities among the $V$ tasks, the graph based multi-task learning formulation can be expressed as follows:
\begin{equation}\label{eq:2}
\min_{W} \sum_{i=1}^V \begin{Vmatrix} X_i \vec{\beta}_i - Y_i \end{Vmatrix}_{F}^{2} + \lambda_1
\begin{Vmatrix} WG\end{Vmatrix}_F^2 +
\lambda_2
\begin{Vmatrix} W\end{Vmatrix}_1
\end{equation}
where $\lambda_1$ and $\lambda_2$ are regularization parameters.
The affinity graph $G$ may be constructed by adding an edge between two tasks that are related (unweighted version), or by using the similarity score between each pair of tasks (weighted version).
To build the graph structure matrix $G$, we first run a multi-task lasso with least squares loss~\cite{tibshirani1996regression} to calculate a basic model.
Then we use this basic model to calculate the correlation coefficients between tasks.
If the correlation coefficient of the two tasks is larger than the correlation threshold, which is set manually, an edge connecting the two tasks will be added.

Note that graph based multi-task learning is much more flexible than mean-regularized multi-task learning as it can incorporate any form of pairwise similarities.
In fact, the mean-regularized multi-task learning can be viewed as a special form of graph based multi-task learning, because it can be achieved by setting graph structure matrix $G$ to connect any arbitrary tasks.
However, the price of flexibility is more parameter tuning and computational complexity.

\subsection{Prediction and Risk Analysis}\label{risk_analysis}

A multi-task learning model trained as in the previous step can be applied to any growing condition input, and produce a predicted yield for any given variety.
Then, we can select the best varieties for each location in the region by striking a tradeoff between the expected yield and growth risk.

The growth risk primarily comes from the unpredictability of weather conditions.
For each location in the ``Region'' dataset, we use its long-term weather data (15 years, from 2001 to 2015) and estimate the yield of each variety using our trained multi-task learning model.
Then, based on such 15 data points, we shall be able to find out the mean and standard deviation of the yield, as if the variety were grown in that location that year.
If our predictive model is fairly accurate, and the weather conditions in the past 15 years were typical, this step will give us an idea about each seed variety's typical performance (i.e., expected yield) and risk for that location, with soil conditions and weather uncertainty accounted for.

To find the best seed variety mix by striking a balance between yield and risk, we employ the asset allocation methodology based on modern portfolio theory (MPT).
As shown in Figure 1, each variety (30 varieties shown) may be represented as a red dot, where the x-axis represents growth risk and the y-axis represents the expected yield.
If we draw a vertical line that corresponds to the maximal tolerable risk, we might find the best mix by carefully selecting the available varieties' weights.
We can then put the performance of this best mix (maximized yield for a given risk) in the plot using a blue dot.
All such blue dots make up a curve known as the ``efficient frontier.''
Any solution on this curve would be optimal at respective risk levels.

\begin{figure}[h]
\centering
\includegraphics[width=0.5\textwidth]{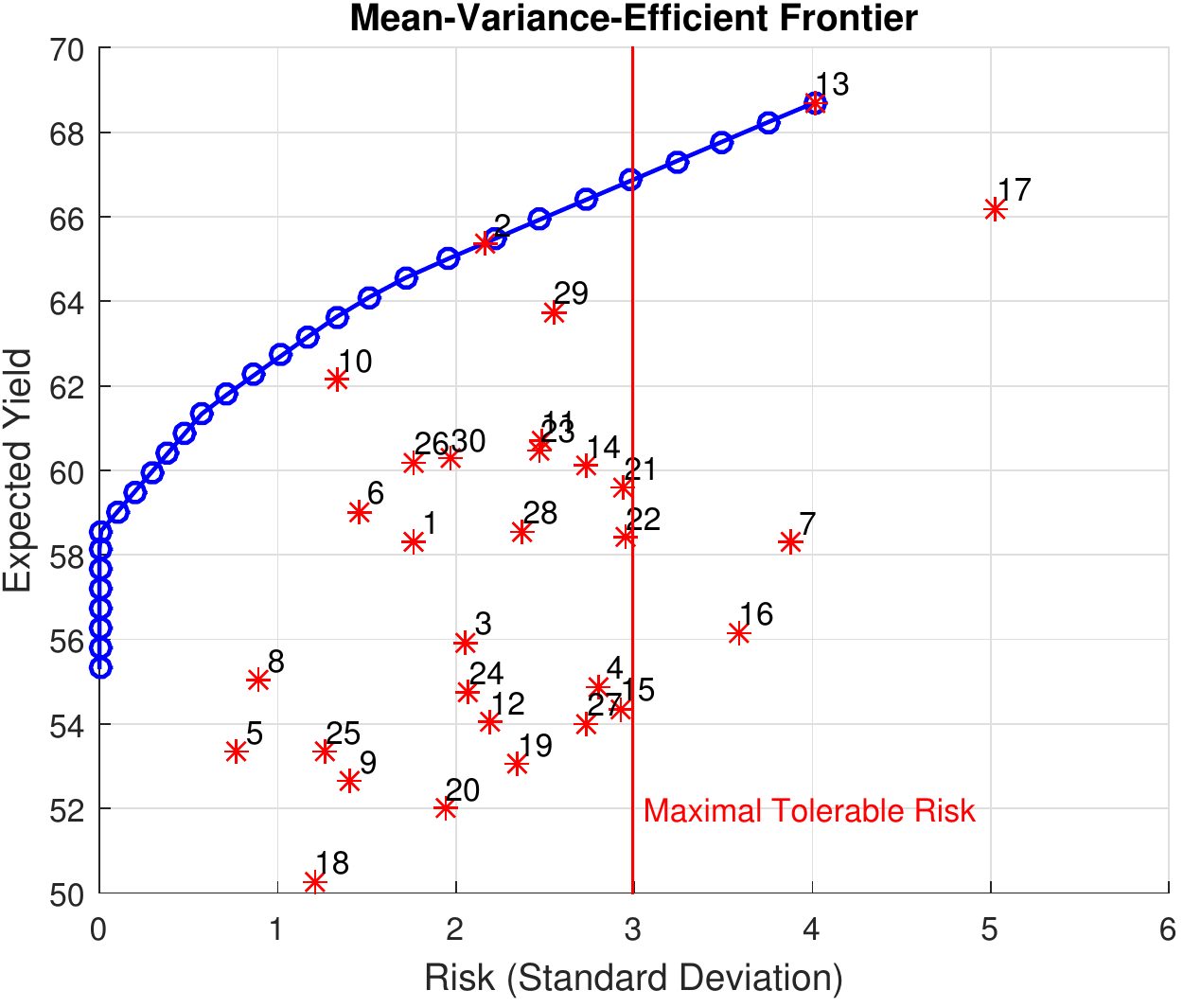}
\caption{MPT efficient frontier}
\label{fig:port}
\end{figure}

The next step is to determine the maximal tolerable risk for each location.
We know that the farmer's risk preferences can be linked to the lands' PI.
Therefore, the maximal tolerable risk can be estimated for each PI level.
In general, the lower PI, the more risk averse.
Without knowing more details about the risk profiles, we choose the risk level for each PI level using a linear scheme.
Specifically, the PI values in the ``Region'' dataset range from $0$ to $18$.
Suppose that the maximal tolerable risk for a location with $PI=0$ is $R_{min}$, and
the maximal tolerable risk for a location with $PI=18$ is $R_{max}$, we will select the maximal tolerable risk for a location with $PI=k$ as
\begin{equation}
R_k = R_{min} + \frac{k}{18}*(R_{max} - R_{min})
\end{equation}
In our experiments, we selected $R_{min} = 0.1$ and $R_{max} = 5.1$.
Then, for each location with $PI=k$, we set the maximal tolerable risk as $R_k$, and use the MPT model to find the location-specific weight vector $\vec{w}_l$ that corresponds to the proportions of each seed variety that should be included in the optimal mix for location $l$.

Formally, for location $l$, suppose that if we only plant variety $i$ next year, the yield will be $\mathcal{Y}_{li}$, which is a random variable (given random weather).
Further suppose its PI value is $k$ (which is easy to find out in the ``Region'' dataset),
then we optimize the following problem:
\begin{equation}\label{eq:optw}
\begin{aligned}
& \underset{\vec{w}_l}{\max}
& & \mathcal{Y}(\vec{w}_l) \equiv \vec{w}_l^T \mathcal{Y}_l, \\
& \text{s.t.}
& & R^2(\vec{w}_l) \equiv \vec{w}_l^T Cov(\mathcal{Y}_l) \vec{w}_l
%\sqrt{\sum\limits_{i=1}^V\sum\limits_{j=1}^Vw_iw_jcov(i,j)}
\quad \leq \quad R^2_k, \\
&&& 0 \leq w_{li} \leq 1, \quad \forall i=1,2,\dots,V; \sum_{i=1}^V w_{li} = 1. \\
\end{aligned}
\end{equation}
where $\mathcal{Y}_l = \left( \mathcal{Y}_{l1}, \mathcal{Y}_{l2}, \dots, \mathcal{Y}_{lV} \right)^T$ is a vector of length $V$ that consists of the (random) yield of each variety at location $l$, and $\vec{w}_l= (w_{l1}, w_{l2}, \dots, w_{lV})^T$ is a vector of length $V$ that corresponds to the proportion of each variety in the optimal mix at location $l$.
$\mathcal{Y}_l$ and $Cov(\mathcal{Y}_l)$ are estimated using our model predicted yield values for all varieties given location $l$'s soil conditions and weather data.
Note that the optimal variety proportions found are location specific.

\subsection{Planning}

Initially, we had considered a bilevel optimization formulation, where the seed company and the farmers make their best decisions simultaneously. However, it took a long time to find the best solution given the high dimensionality of our problem. Therefore, we developed a simple and intuitive heuristic method, as presented below.

\subsubsection{Popular Seed Ranking}\label{sec:ranking}

In this step, we narrow down to a small number of seed varieties by total demand.
Given the location-specific optimal weights estimated from Formulation~(\ref{eq:optw}), we can aggregate the total demand for seed variety $i$ as:

\begin{equation}\label{eq:demand}
d_i = \sum_{l=1}^N a_l w_{li}, \quad i = 1,2,\dots,V
\end{equation}

where $a_l$ is the area size for growing seeds at location $l$, $w_{li}$ is the optimized proportion of variety $i$ at location $l$, and $N$ is the total number of locations.
Then we pick the top $M$ ($M\geq 5$) most popular seed varieties as the final candidates.
In this way, we can reduce the number of candidates for fast optimization.

\subsubsection{Weight Optimization for Top 5 Seed Varieties}\label{sec:reweighting}

The eventual goal of this study is to propose the proportions of up to five seed varieties.
By providing these five varieties only, farmers cannot always achieve their personalized optimal mix as suggested in Formulation~(\ref{eq:optw}).
Therefore, in this step, we shall re-adjust seed seller's stocking proportions by minimizing the total gap of each location's optimality.

More specifically, suppose that farmers will need to optimize seed mix within the $M$ offered by the seller. As illustrated in Figure~\ref{fig:objfuction}, they can re-balance the weights and produce a new ``efficient frontier'' using just these $M$ varieties.
Cutting at the same maximal tolerable risk level, the difference in expected yield from the old and new efficient frontiers would represent the ``suboptimality gap.''
We reuse Formulation~(\ref{eq:optw}) to adjust the final proportions of the $M$ varieties.

\begin{figure}[h]
\centering
\includegraphics[width=0.6\textwidth]{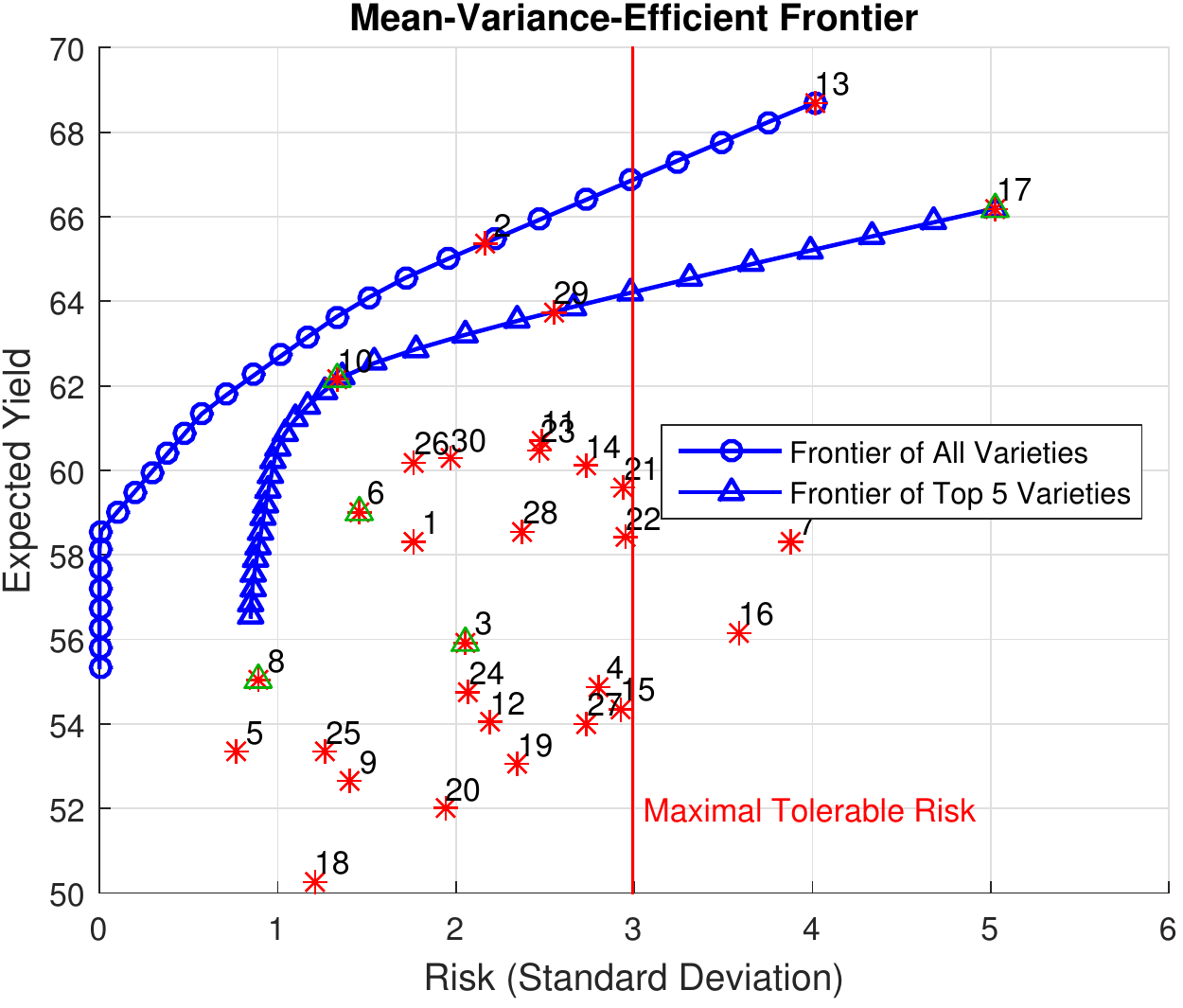}
\caption{Objective function definition}
\label{fig:objfuction}
\end{figure}

\section{Implementation}\label{design}

This section presents our implementation details, including data preprocessing, multi-task learning model training and evaluation, and parameter tuning.

\subsection{Data Preprocessing}

For different attributes, value ranges vary wildly. For example, the solar radiation ranges from 900,000 to 1,300,000, while the pH ranges from 5.0 to 9.0. To make the data more reasonable to feed multi-task learning models, we rescale the values of all weather- and soil-related attributes in both ``Experiment'' and ``Region'' datasets using min-max normalization.

\subsection{Training \& Testing Models}

We use the MALSAR multi-task learning packages published by~\cite{zhou2012mutal} to train the mean-regularized multi-task model and the graph-based multi-task model. We randomly split the ``Experiment'' dataset into 80\% for training and 20\% for testing.
On the training dataset, a five-fold cross validation with root mean squared error (RMSE) is used to estimate the regularization parameters for both models. A commonly used RMSE in multi-task regression is expressed as follows:

\begin{equation}
RMSE = \frac{\sum_{i=1}^{t}\sqrt{\sum_{j=1}^{n_i}(X_{i,j}*W_i-Y_{i,j})^2}*n_i
}{\sum_{i=1}^{t}n_i}
\end{equation}

where $t$ is the total number of tasks, $n_i$ is the number of observations for task $i$,
where $W_i$ is the model of the task $i$, $X_{i,j}$ is the $j$th observation data for the task $i$, $Y_{i,j}$ is the $j$th target yield for the task $i$.
Then we use the best parameters obtained to build models and evaluate their performance on the testing dataset using RMSE.

\subsection{Tuning Regularization Parameters}\label{eval_para}

For the mean-regularized multi-task model, only one parameter $\lambda$ (see Equation \ref{eq:1}) is required to estimate. We iterate $\lambda \in \{10^{-2}, 10^{-1}...10^{2}\}$ and select optimal $\lambda = 0.01$ which has a lowest validation RMSE.

For the graph based multi-task model, the parameters $\lambda_1$ and $\lambda_2$ (see Equation \ref{eq:2}) need to be tuned. In addition, to create the graph structure matrix $G$ in Equation \ref{eq:2}, a lasso regularizing parameter $\lambda_L$ and a correlation threshold $t$ among tasks also need to be determined.
If the correlation of two tasks is larger than the threshold, an edge indicating that the two tasks are related is added between the two tasks. More details about $\lambda_L$ and $t$ can be found in MALSAR user's manual by~\cite{zhou2012mutal}. We iterate the combination of \{$\lambda_1$,$\lambda_2$,$\lambda_L$,$t$\} where $\lambda_1 \in \{10^{-2}, 10^{-1}...10^{2}\}, \lambda_2 \in \{10^{-2}, 10^{-1}...10^{2}\}, \lambda_L \in \{10^{-2}, 10^{-1}...10^{2}\}, t \in \{0.1, 0.2...1.0\} $ and find the optimal $\lambda_1 = 0.1$,$\lambda_2 =0.1$, $\lambda_L =0.01$ and $t=0.9$.

\section{Results and Evaluations}\label{evaluation}

In this section, we evaluate the performance of mean-regularized and graph based multi-task models.
Also, we demonstrate the top five seed varieties and their corresponding proportions for stocking. 

\subsection{Performance of Mean-Regularized and Graph-based Multi-Task Models}

Using the optimal parameters above, we evaluate the performance of graph based multi-task model and mean-regularized multi-task model on the testing dataset.
We find that the graph-based multi-task model outperforms the mean-regularized multi-task model slightly with a RMSE of $10.008$ and $10.023$ respectively.
So we use the mean-regularized multi-task model to estimate the yield of locations in ``Region'' dataset.
Besides, we demonstrate the two multi-task models outperform Naive Bayes significantly, as shown in Figure~\ref{fig:naivebayes}.

\begin{figure}
    \centering
    \begin{minipage}{.5\textwidth}
        \centering
        \includegraphics[width=\textwidth]{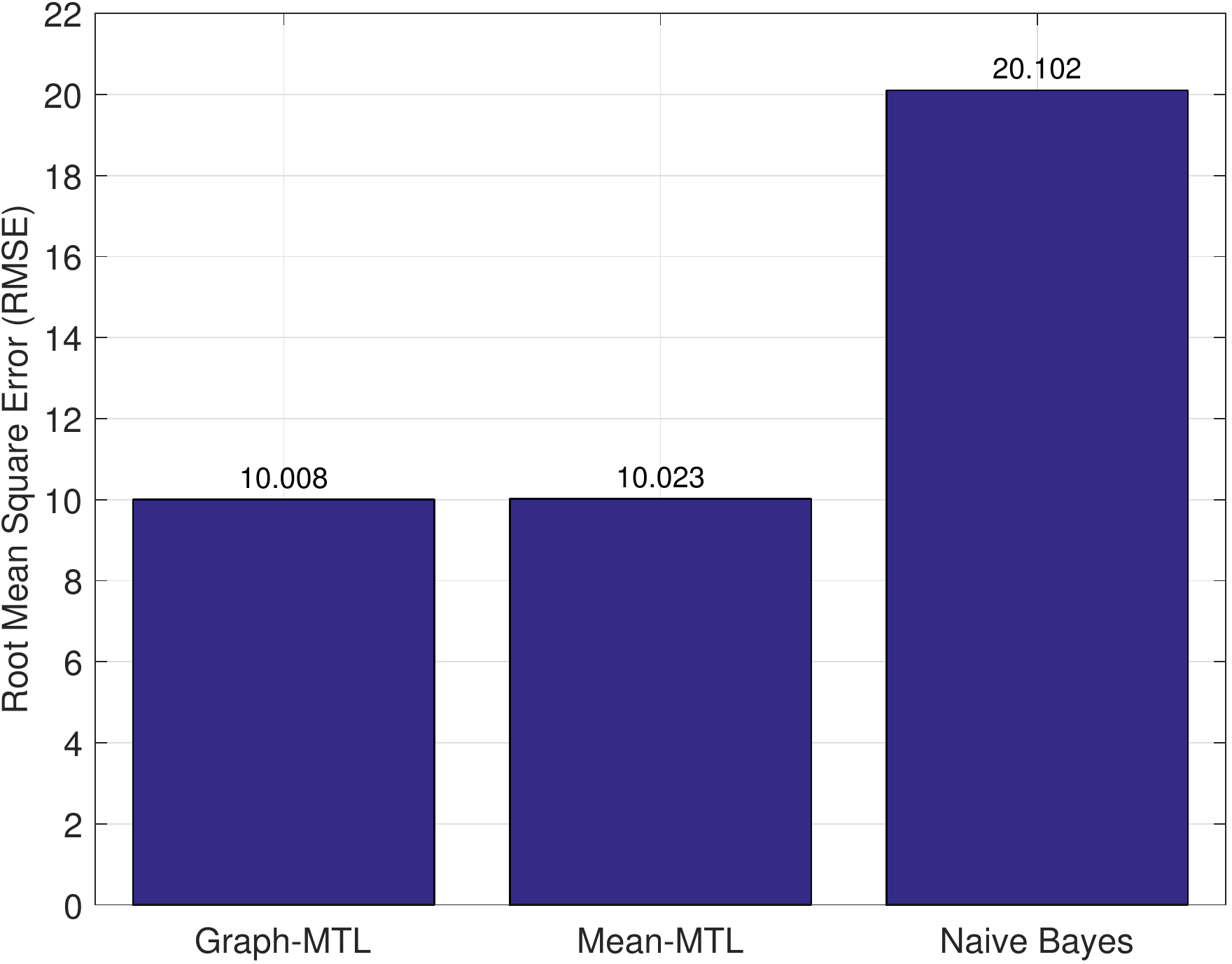}
        \caption{Performance comparison with naive bayes model}
        \label{fig:naivebayes}
    \end{minipage}%
    \begin{minipage}{.47\textwidth}
        \centering
        \includegraphics[width=\textwidth]{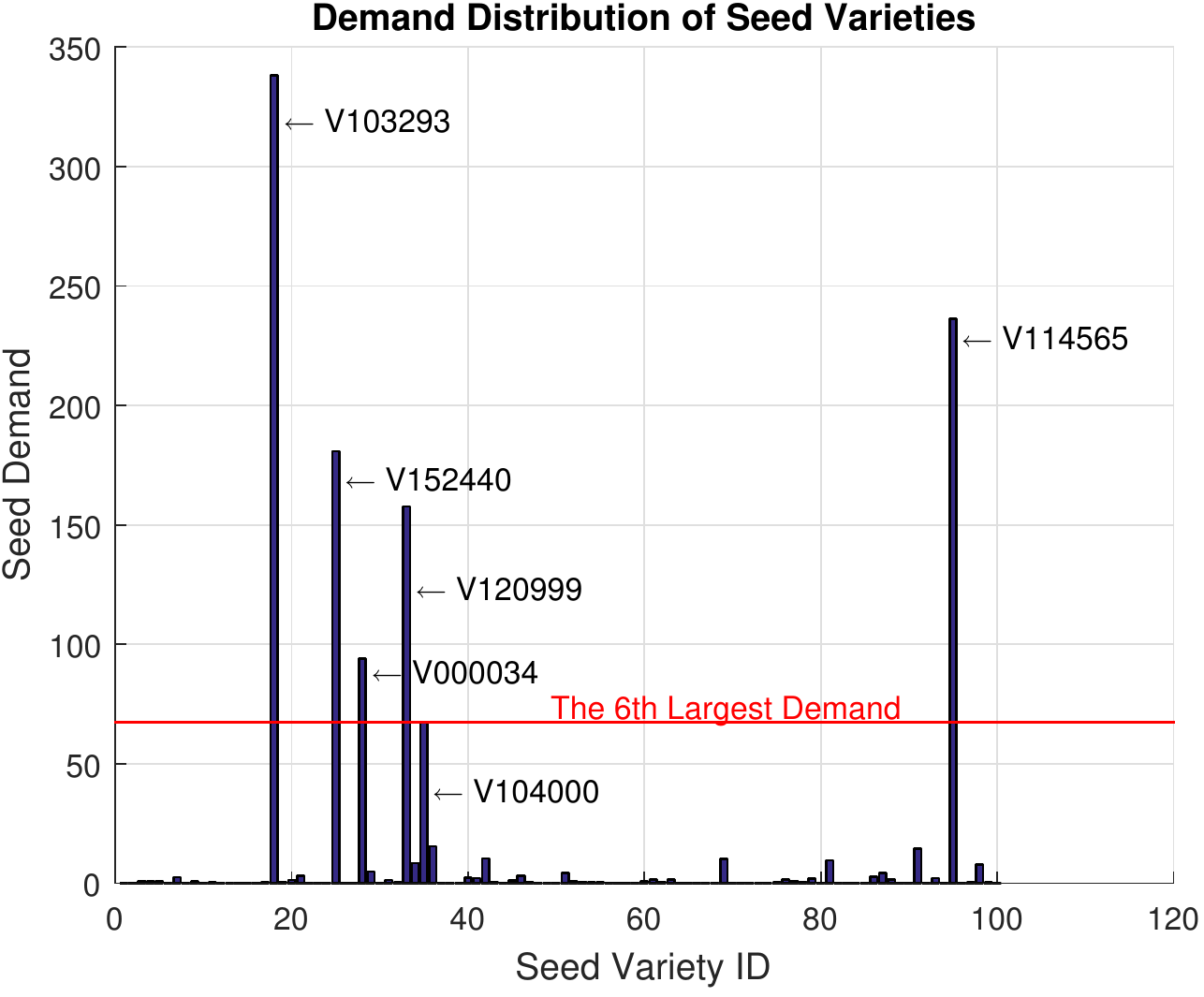}
        \caption{Seed demand of different varieties}
        \label{fig:demand}
    \end{minipage}
\end{figure}

\subsection{Selection of Top Five Candidate Varieties}

We use the approach presented in Section~\ref{sec:ranking} to calculate the demand for each variety, as visualized in Figure~\ref{fig:demand}.
We can see that the top four (or top six) seed varieties by demand exceed other varieties by a large margin.

\subsection{Proportion of Selected Seed Varieties}

Using the approach presented in Section~\ref{sec:reweighting}, we optimize seed mix among these top seeds for each location separately, and then aggregate all the 5 seed varieties together.
The aggregated proportions are shown in Figure~\ref{fig:pie}.
Since one of the varieties was below 10\%, we dropped it from our final solution, and redistributed the proportions among the remaining four varieties.
Therefore, our recommendation is to stock (i) 28\% of Variety V103293,
(ii) 20\% of Variety V114565,
(iii) 38\%  of Variety V152440, and
(iv) 14\% of Variety V120999.

\begin{figure}[h]
\centering
\includegraphics[width=0.4\textwidth]{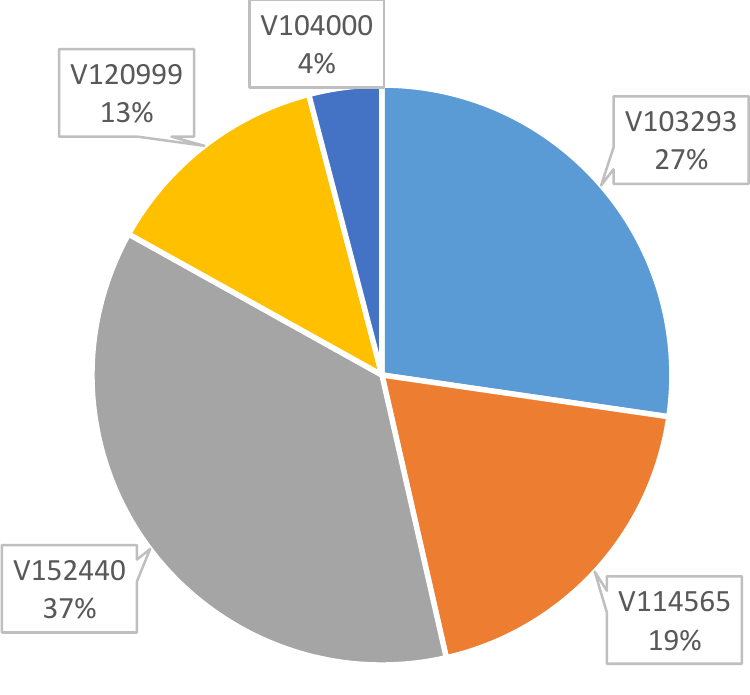}
\caption{Proportion of seed varieties}
\label{fig:pie}
\end{figure}

\section{Conclusions \& Future Work}\label{conclusion}

In this paper, we propose a framework to solve the socially beneficial problem of selecting seed variety combinations for stocking. To be specific, we first use multi-task learning techniques to estimate yields of different locations by leveraging the commonalities as well as the uniqueness among different seed varieties. Then we determine the best mix of seeds for each location by seeking a tradeoff between yield and risk. Then we pick the top five varieties based on the aggregated best mix of each location. Finally, we re-balance each location's yield and risk by only growing a mix or a single of the top five varieties.

More work will be done for tuning parameters in a smaller granularity for both mean-regularized and graph based multi-task models. Other multi-task learning variations, like Robust MTL, Relaxed ASO, and Fused Sparse Group Lasso will be applied and be evaluated. This work may make the yield predictions more accurate.

\bibliographystyle{unsrt}
\bibliography{reference}

\end{document}